\documentclass{siam-dr-article}

\usepackage{amssymb,amsmath,dsfont}
\usepackage{bm}
\usepackage{comment}

\newcommand{\ie}{\textit{i.e.}}

\newcommand{\quartertab}{\hspace*{.5em}}
\newcommand{\eighthtab}{\hspace*{.25em}}

\newtheorem{definition}{Definition}

\DeclareMathOperator*{\argmax}{arg\,max}

\title{Detection of Abnormal Input-Output Associations}

\author{Charmgil Hong, Siqi Liu, Milos Hauskrecht}

\usehyperref 

\begin{document}
\maketitle

\section*{Summary}
We study a novel outlier detection problem that aims to identify abnormal input-output associations in data, whose instances consist of multi-dimensional input (context) and output (responses) pairs.
We present our approach that works by analyzing data in the conditional (input--output) relation space, captured by a decomposable probabilistic model.
Experimental results demonstrate the ability of our approach in identifying multivariate conditional outliers.

\section{Background}
Outlier detection is a data analysis task that aims to find atypical behaviors, unusual outcomes, erroneous readings or annotations in data. 
It has been an active research topic in data mining community, and is frequently adopted in various applications to identify rare and interesting data patterns, which may be associated with beneficial or malicious events \cite{garcia:2009,Wang:2010:ICICTA,Hauskrecht:2016:JBI}.
%
Despite an extensive research, however, the majority of existing approaches are developed to detect \textit{unconditional} outliers that take place in the joint space of all data attributes.
Such methods may not work well when one wants to identify \textit{conditional} (contextual) outliers that reflect unusual responses for a given set of contextual attributes. 
Since conditional outliers reveal themselves given the context or properties of data instances, an application of unconditional outlier detection methods may lead to incorrect results.
For example, provided that we want to identify highly unusual image tags in a collection of annotated images, applying unconditional detection methods to the joint (\textit{image, tags}) space may lead to images with rare themes to be falsely identified as outliers, due to the scarcity of these themes in the dataset (false positives). 
Similarly, unusual annotations on the images with frequent themes may not be discovered as the difference between the outlier and normal instances are only marginal, {especially when the dimensionality of the images dominates the attribute space} (false negative).
%

This work studies \textit{multivariate conditional outlier detection} \cite{Hong:2015:AAAI} that formalizes and addresses the above special case of outlier detection, which concerns irregular associations of input and output patterns in data.
Specifically, we consider problems where each instance in dataset $\mathcal{D}\!=\!\{\mathbf{x}^{(n)}, \mathbf{y}^{(n)}\}_{n=1}^N$ consists of an $m$-dimensional continuous input vector $\mathbf{x}^{(n)}\!=\!(x_1^{(n)}, ..., x_m^{(n)})$ and a $d$-dimensional binary output vector $\mathbf{y}^{(n)}\!=\!(y_1^{(n)}, ..., y_d^{(n)})$.
Our goal is to \textit{detect unusual responses in $\mathbf{Y}$ given context $\mathbf{X}$}.
The fundamental issues in identifying multivariate conditional outliers are how to take into account the {contextual dependences between output $\mathbf{Y}$ and their input $\mathbf{X}$}, as well as the {mutual dependences among $\mathbf{Y}$}. 

In tackling such problems, mainly two approaches are suggested in the literature.
One way is to apply an unconditional detection method only to the output space where, by definition, outliers may occur \cite{aggarwal:2013:book}.
Another is to adopt subspace analysis methods for outlier detection that find lower-dimensional surrogate representations of data that are effective for outlier detection \cite{Lazarevic:2005:KDD,keller:2012:icde}.
However, both approaches do not fully consider the underlying conditional dependence relations and thereby cannot properly perform the identification of multivariate conditional outliers.

This paper presents a novel subspace approach for multivariate conditional outlier detection that considers both the contextual- and inter-dependences among data attributes. 
In particular, we develop a model-based approach that first builds a decomposable probabilistic data model of conditional joint $P(\mathbf{Y}|\mathbf{X})$; 
and then applies the model to project data to a lower-dimensional conditional probability space, where the data instances are assessed for outliers.
Our approach hence consists of two phase: (1) multi-dimensional model building and (2) model-based transformation.
Below we discuss each phase in detail.

\section{Approach}

\noindent
\textit{\textbf{Probabilistic Multi-dimensional Output Models}}\\
Our outlier detection approach works by analyzing data instances come in input-output pairs with a statistical model representing the conditional joint distribution $P(\mathbf{Y}|\mathbf{X})$. 
Accordingly, our objective in the first phase is to efficiently build a compact representation of complex input--output relations.
A direct learning of the conditional joint from data, however, is often very expensive or even infeasible as the number of possible output combinations grows exponentially with $d$.
%
To avoid such a high modeling cost yet achieve an accurate data representation for outlier detection, we decompose the conditional joint into a product of conditional univariates using the chain rule of probability:
$P(Y_1,...,Y_d | \mathbf{X}) = \prod_{i=1}^d P(Y_i|\mathbf{X},\mathbf{Y}_{\boldsymbol{\pi}(i)})$, 
where $\mathbf{Y}_{\boldsymbol{\pi}(i)}$ denotes the parents of $Y_i$ \cite{Hong:2015:SDM}.
This decomposition lets us represent $P(\mathbf{Y}|\mathbf{X})$ by simply specifying each univariate conditional factor, $P(Y_i|\mathbf{X},\mathbf{Y}_{\boldsymbol{\pi}(i)})$.
We parameterize our data representation $\mathcal{M}$ of $\mathcal{D}$ as:\footnote{In this work, we use the l-2 regularized logistic regression.}

\vspace{-1.5em}
\begin{align}
	\label{eq:param}
	\theta_{\mathcal{M}(i)} = \argmax_\theta \sum_{n=1}^N \log P(y_i^{(n)}|\mathbf{x}^{(n)},\mathbf{Y}_{\boldsymbol{\pi}(i)}; \theta)
\end{align}
\vspace{-1.15em}

\noindent
where $\theta_{\mathcal{M}(i)} : i = 1, ..., d$ denotes the parameters of the probabilistic model for the $i$-th output dimension.
\\

\vspace{-.5em}
\noindent
\textit{\textbf{Transformation for Outlier Detection}}\\
In the second phase, our objective is to assess the abnormality of each input-output association using our data representation $\mathcal{M}$.
Let us apply $\mathcal{M}$ to estimate the conditional probabilities of observed outputs. 
For notational convenience, we introduce an auxiliary vector $\boldsymbol{\rho}\!=\!(\rho_1, ..., \rho_d)$ that each element $\rho_i$ 
 is quantized by a probabilistic estimation process, which is formalized by unleashing the product in the chain rule: 

\vspace{-1.5em}
\begin{align}
\mathcal{M}: \eighthtab(\mathbf{x}^{(n)},\mathbf{y}^{(n)}) \quartertab\rightarrow\quartertab\boldsymbol{\rho}^{(n)} 
				& = (\rho_1^{(n)}, ..., \rho_d^{(n)} ) 
\end{align}
\vspace{-1.5em}

\noindent
where $\rho_i^{(n)} = \widetilde{P}(y_i^{(n)}|\mathbf{x}^{(n)},\mathbf{y}_{\boldsymbol{\pi}(i)}^{(n)}; \theta_{\mathcal{M}(i)})$.
Accordingly, the space of $\boldsymbol{\rho}$ is projecting a normalized confidence level (\ie, conditional probability estimate) of each observation $(\mathbf{x}^{(n)},\mathbf{y}^{(n)})$ across individual output dimensions. 
Now we consider observed conditional probability estimates $\boldsymbol{\rho}$ as low-dimensional proxies of the original instances, and further hypothesize that multivariate conditional outliers could be effectively detected in this proxy space. 
Next section shows how this approach improves the multivariate conditional outlier detection results.

\section{Evaluation and Discussion}

We demonstrate the performance of our outlier detection approach in combination with two widely used multivariate outlier detection methods: \textit{local outlier factor} (LOF) \cite{breunig:2000} and \textit{one-class SVM} (OCS) \cite{scholkopf:1999:NIPS}.
We compare our results with that of two conventional outlier detection strategies: applying the base detection methods either to the joint space of input and output (JOINT), or only to the output space (OUT);
and that of two state-of-the-art subspace analysis techniques developed for outlier detection: \textit{feature bagging} (FB) \cite{Lazarevic:2005:KDD} and \textit{high contrast subspaces} (HICS) \cite{keller:2012:icde}.

For data, we use two scene image datasets, annotated with semantic tags \cite{Boutell:2004:PR,Zhang:2007:PR}.
The properties of the datasets are listed in Table \ref{table:datasets}.\footnote{For example, \textit{Dataset \#1} contains 2,407 images represented by feature vectors of length 294. Each image comes with a binary vector of length 6 that indicates whether the image has been annotated with the tags listed in the table.}
On each test, we perturbed 1\% of the data instances by randomly flipping \textit{one} output variable (\ie, representing images with an unusual tag), and evaluated how the methods identify them.

\begin{figure}[t]
\centering
	\includegraphics[width=0.405\textwidth]{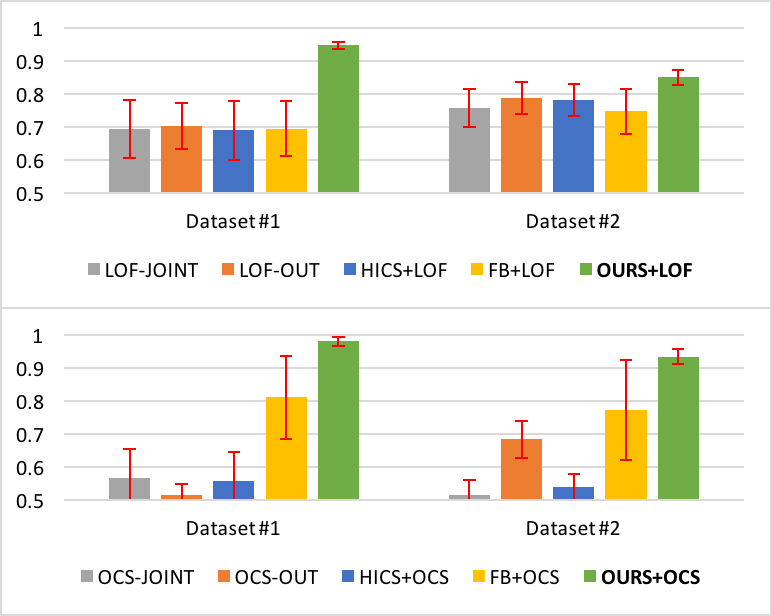}
	\vspace{-1.05em}
	\caption{Performance comparison in AUC.}
	\label{fig:results}
	\vspace{-.65em}
\end{figure}

Figure \ref{fig:results} shows the results in terms of the area under the receiver operating characteristic curve (AUC). 
Each bar plots the mean and standard error (red line) in AUC after ten repeats.
Through the results, we verify that our proposed approach (OURS+LOF and OURS+OCS) not only improves the performance of the base methods (LOF-JOINT and OCS-JOINT), but also consistently outperforms the existing subspace outlier detection approaches.



To conclude, we summarized our model-based approach for the multivariate conditional outlier detection problem.
Experimental results support the advantages of our approach in addressing the problem.

\begin{table}[t]
\begin{center}
    {\fontsize{6.5}{7.5}\rm
	\begin{tabular}{ c  c  c  c  c  c  c }
		\hline
		\textbf{Dataset} &  \textsc{$N$} &  \textsc{$m$} &  \textsc{$d$} &  \textbf{Tags}\\
		\hline\\[-.85em]
		Dataset \#1        & 2,407          & 294            & 6              & \begin{tabular}[c]{@{}c@{}} beach, sunset, foliage, field, \\ mountains, urban \end{tabular} \\
		\\[-.75em]
Dataset \#2        & 2,000          & 135            & 5              & \begin{tabular}[c]{@{}c@{}} desert, mountains, sea, \\ sunset, trees \end{tabular}
		\\
		\hline
	\end{tabular}
	}
	\caption{ Datasets ($N$: number of instances, $m$: input dimensionality, $d$: output dimensionality).}
	\label{table:datasets}
\end{center}
\end{table}


\bibliographystyle{abbrv}
\bibliography{dr17}

\end{document}